\documentclass[11pt,twocolumn]{article}

\usepackage[numbers]{natbib}

\usepackage[vlined,linesnumbered,ruled,resetcount]{algorithm2e}
\usepackage[colorlinks,linkcolor=magenta,filecolor=blue,citecolor=blue,urlcolor=blue]{hyperref}%
\usepackage[top=1in, left=1in, right=1in, bottom=1in]{geometry}
\usepackage{pifont}

\usepackage{authblk}
\usepackage{enumitem}

\usepackage{times}
\usepackage{latexsym}

\usepackage{graphicx}   
\usepackage{caption}    
\usepackage{subcaption} 

\usepackage[T1]{fontenc}

\usepackage[utf8]{inputenc}

\usepackage{microtype}

\usepackage{inconsolata}

\usepackage{graphicx}
\usepackage{enumitem}

\usepackage{array}
\usepackage{booktabs}
\usepackage{multirow}
\usepackage{amssymb}

\usepackage{xcolor}         
\usepackage{multirow}
\usepackage{diagbox}
\usepackage{pifont}

\definecolor{redx}{RGB}{180,0,0}
\definecolor{greenx}{RGB}{0,180,0}

\newcommand{\redxmark}{\color{redx}\ding{55}}
\newcommand{\greencmark}{\color{greenx}\ding{51}}
\definecolor{redx}{RGB}{180,0,0}
\definecolor{greenx}{RGB}{0,180,0}

\usepackage{threeparttable,booktabs}
\usepackage{amsmath}

\makeatletter
\newcommand*{\RN}[1]{\expandafter\@slowromancap\romannumeral #1@}
\makeatother

\makeatletter
\newcommand{\printfnsymbol}[1]{%
  \textsuperscript{\@fnsymbol{#1}}%
}
\makeatother

\title{\huge{Federated Large Language Models:\\ Current Progress and Future Directions}}

\author{\normalsize{
 {Yuhang Yao\textsuperscript{1}},
 {Jianyi Zhang\textsuperscript{2}},
 {Junda Wu\textsuperscript{3}},
 {Chengkai Huang\textsuperscript{4}},
 {Yu Xia\textsuperscript{3}},
 {Tong Yu\textsuperscript{5}},
 {Ruiyi Zhang\textsuperscript{5}},\newline
 {Sungchul Kim\textsuperscript{5}},
 {Ryan Rossi\textsuperscript{5}},
 {Ang Li\textsuperscript{6}},
 {Lina Yao\textsuperscript{4,7}},
 {Julian McAuley\textsuperscript{3}},
 {Yiran Chen\textsuperscript{2}},
 {Carlee Joe-Wong\textsuperscript{1}}
\\
 \textsuperscript{1}Carnegie Mellon University,
 \textsuperscript{2}Duke University,
 \textsuperscript{3}University of California San Diego,\\
 \textsuperscript{4}The University of New South Wales,
 \textsuperscript{5}Adobe Research, \\
 \textsuperscript{6}University of Maryland College Park,
 \textsuperscript{7}CSIRO's Data61\\
   yuhangya@alumni.cmu.edu
}}

\begin{document}
\date{}
\maketitle
\begin{abstract}
Large language models are rapidly gaining popularity and have been widely adopted in real-world applications. While the quality of training data is essential, privacy concerns arise during data collection. Federated learning offers a solution by allowing multiple clients to collaboratively train LLMs without sharing local data. However, FL introduces new challenges, such as model convergence issues due to heterogeneous data and high communication costs. A comprehensive study is required to address these challenges and guide future research. This paper surveys Federated learning for LLMs (FedLLM), highlighting recent advances and future directions. We focus on two key aspects: fine-tuning and prompt learning in a federated setting, discussing existing work and associated research challenges. We finally propose potential directions for federated LLMs, including pre-training, federated agents, and LLMs for federated learning.
\end{abstract}

\section{Introduction}\label{sec:intro}

Large Language Models (LLMs) have transformed artificial intelligence by excelling in understanding and generating human-like text, advancing areas like natural language processing and code generation. However, their training is computationally expensive, requiring large datasets and high-performance computing, which poses challenges for many organizations, particularly around data privacy. In fields like healthcare, finance, and legal services, sharing sensitive data for centralized training is risky due to regulatory concerns and potential data breaches. Federated learning offers a solution by allowing decentralized model training, where participants collaborate without sharing raw data. Instead, only model updates are exchanged, reducing privacy risks while benefiting from a collective, diverse dataset.

\begin{table*}[ht]
\caption{Comparison with Existing Surveys on Federated Large Language Models.}
\vspace{-1em}
\centering
\resizebox{\textwidth}{!}{
\begin{tabular}{|c|c|c|c|c|c|}
\hline
 & \cite{chen2023federated} & \cite{yu2023federated} & \cite{woisetschlager2024survey} & \cite{zhuang2023foundation} & Ours \\ \hline
Security \& Privacy & \greencmark & \greencmark & \redxmark & \greencmark & \greencmark \\ \hline
Communication \& Computation Efficiency  & \redxmark & \redxmark & \greencmark & \greencmark & \greencmark \\ \hline
Prompt Tuning \& PEFT Methods & \redxmark & \redxmark & \greencmark & \redxmark & \greencmark \\ \hline
Real World Applications & \redxmark & \redxmark & \redxmark & \redxmark & \greencmark \\ \hline
Emerging Directions (Agents / Pre-training) & \redxmark & \redxmark & \redxmark & \redxmark & \greencmark \\ \hline
\end{tabular}}
\label{tab:former_work}
\end{table*}

The integration of FL with LLMs introduces unique challenges and opportunities. The sheer size of LLMs and the complexity of their training processes require careful consideration of communication overhead, computational efficiency, and convergence stability. Researchers have started to address these issues, leading to a growing body of literature that explores various aspects of Federated Learning for LLMs (FedLLM). 
Several surveys have attempted to capture the evolving landscape of FedLLM research. For instance,~\cite{chen2023federated} and~\cite{yu2023federated} focus on privacy protection mechanisms within FL frameworks for LLMs. ~\cite{woisetschlager2024survey} take a different approach by concentrating on communication and computation efficiency. They analyze strategies to reduce the communication overhead between clients and the central server, such as model compression, sparsification, and quantization. ~\cite{zhuang2023foundation} offers insights into the future of foundation models, including LLMs, but lacks a comprehensive review of recent papers on federated learning. 


Recognizing the gaps in existing literature, as shown in Table~\ref{tab:former_work}, this survey aims to provide a comprehensive and up-to-date overview of recent advances in federated learning for LLMs. Our goal is to serve as a valuable reference for researchers and practitioners interested in this interdisciplinary domain. We systematically review the latest methodologies, highlight their contributions, and discuss how they address the inherent challenges of FL with LLM training. 

In this survey, we first review recent advances in federated fine-tuning and federated prompt learning for large language models. We then analyze key challenges, including efficiency, personalization, privacy, and security, and summarize representative real-world applications. Finally, we discuss emerging directions such as federated pre-training and federated AI agents.

\section{Federated Fine-Tuning}
Fine-tuning is one of the most important directions of FedLLMs. Traditional topics in FL are still relevant but introduce new challenges in the context of FedLLMs, as shown in Table~\ref{tab:topic_fine_tuning}. Efficiency plays a key role in reducing the high cost of training. Personalization enables the creation of AI assistants tailored to individual users. Privacy and security become more critical in the context of LLMs, given the sensitive nature of the data involved. Instruction-tuning and value alignment enhance model accuracy and mitigate risks in applying LLMs.

\begin{table*}[ht]
\caption{Overview of Topics in Federated Fine-Tuning.}
\vspace{-1em}
\resizebox{\textwidth}{!}{
\begin{tabular}{|c|c|c|}
\hline
\textbf{Area}                        & \textbf{Topic}                           & \textbf{Approaches}                                                                                \\ \hline
\multirow{2}{*}{Heterogeneity}        & Data Heterogeneity                      & \begin{tabular}[c]{@{}c@{}}FedDAT~\cite{chen2023feddat}, RaFFM~\cite{yu2023bridging}, FedKC~\cite{wang2022fedkc}, \\ ~\cite{wang2023federated}, FEDSA~\cite{zhangenhancing}\end{tabular}                                                                                                                                        \\ \cline{2-3} 
                                      & Model Heterogeneity & \begin{tabular}[c]{@{}c@{}}FedLoRA~\cite{yi2023fedlora}, HETLORA~\cite{cho2024heterogeneous}, FlexLoRA~\cite{bai2024federated}, \\ FedPCL~\cite{tan2022federated}, FedBRB~\cite{xu2024fedbrb}\end{tabular}                               \\ \hline
\multirow{4}{*}{Privacy and Security} & Security and Robustness                 & ~\cite{lu2023exploring}, ~\cite{zhao2023privacy},~\cite{wu2024vulnerabilities}, ~\cite{han2023fedsecurity}                                                                                                                                                                                                                                      \\ \cline{2-3} 
                                      & Privacy                                 & FedPIT~\cite{zhang2024fedpit}, FFA-LoRA~\cite{sun2024improving}                                    \\ \cline{2-3} 
                                      & Attacks                                 & ~\cite{wu2024vulnerabilities}, ~\cite{gupta2022recovering}, Decepticon~\cite{fowl2022decepticons}, ~\cite{li2023unveiling}                                                                                                                                                                                                                      \\ \cline{2-3} 
                                      & Defense                                 & FEDML-HE~\cite{jin2023fedml}, ~\cite{yu2023federated}, FedBiOT~\cite{wu2023fedbiot}                                                                                                                                                                                                                                                                                   \\ \hline
\multirow{3}{*}{Efficiency}           & Training Efficiency                     & Grouper~\cite{charles2024towards}, FedYolo~\cite{zhang2023fedyolo}, FedTune~\cite{chen2022fedtune}                                                                                                                                                                                                                                                                    \\ \cline{2-3} 
                                      & Communication Efficiency                & CEFHRI~\cite{khalid2023cefhri}, FedKSeed~\cite{qin2023federated}, FedRDMA~\cite{zhang2024fedrdma}                                                                                                                                                                                                                                                                     \\ \cline{2-3} 
                                      & Parameter Efficiency                    & \begin{tabular}[c]{@{}c@{}}~\cite{sun2022exploring}, FedPETuning~\cite{zhang2023fedpetuning}, ~\cite{wang2024personalized}, \\ SLoRA~\cite{babakniya2023slora}, FFA-LoRA~\cite{sun2023improving}, \\ FlexLoRA~\cite{bai2024federated}, LP-FL~\cite{jiang2023low}, FedBiOT~\cite{wu2024fedbiot}\end{tabular} \\ \hline
\multirow{5}{*}{Frameworks}           & Cross-Silo                              & FedRDMA~\cite{zhang2024fedrdma}, CROSSLM~\cite{deng2023mutual}                                                                                                                                                                                                                                                                                                                              \\ \cline{2-3} 
                                      & Cross-Device                            & ~\cite{wang2023can}, FwdLLM~\cite{xu2023federated}, ~\cite{woisetschlager2023federated}                                                                                                                                                                                                                                                                               \\ \cline{2-3} 
                                      & Decentralized Training                  & OpenFedLLM~\cite{ye2024openfedllm}, ~\cite{tang2023fusionai}, ~\cite{yuan2022decentralized}                                                                                                                                                                                                                                                                           \\ \cline{2-3} 
                                      & Blackbox and Transfer Learning          & Fed-BBPL ~\cite{lin2023efficient}, ZooPFL ~\cite{lu2023zoopfl}, ~\cite{kang2023grounding}                                                                                                                                                                                                                                                                                  \\ \cline{2-3} 
                                      & Instruction Tuning                      & FederatedScope-LLM~\cite{kuang2023federatedscope}, FedIT ~\cite{zhang2023towards}                                                                                                                                                                                                                                                                                                           \\ \hline
\multirow{3}{*}{Evaluation}           & Datasets                                & FederatedScope-LLM~\cite{kuang2023federatedscope}, OpenFedLLM~\cite{ye2024openfedllm}                                                                                                                                                                                                                                                                                                       \\ \cline{2-3} 
                                      & Benchmarks                              & FedLLM-Bench ~\cite{ye2024fedllm}, Fedmlsecurity~\cite{han2023fedmlsecurity}, Profit~\cite{collins2023profit}                                                                                                                                                                                                                                                               \\ \cline{2-3} 
                                      & Convergence Analysis                    & FedPEAT~\cite{chua2023fedpeat}, FedMeZO~\cite{ling2024convergence}                                                                                                                                                                                                                                                                                                                          \\ \hline
\end{tabular}
}
\label{tab:topic_fine_tuning}
\end{table*}

\subsection{Heterogeneity}
\textbf{Data Heterogeneity}
To address data heterogeneity for various vision-language tasks in federated multimodal fine-tuning, FedDAT~\cite{chen2023feddat} proposes a Dual-Adapter Teacher to regularize clients' local updates and Mutual Knowledge Distillation for knowledge transfer.
RaFFM~\cite{yu2023bridging} proposes a novel model compression method to accommodate heterogeneous training data.~\cite{wang2022fedkc} focuses on addressing the challenges of data heterogeneity in multilingual federated natural language understanding by proposing a FedKC module that exchanges knowledge across clients without sharing raw data.~\cite{wang2023federated} proposes a distillation method from heterogeneous tag sets to achieve better domain adaptation.
In addition to transferring multimodal knowledge in FL, M$^2$FEDSA~\cite{zhangenhancing} proposes split learning to realize modularized decomposition and enables unimodal local models to be complementarily aggregated. 

\noindent\textbf{Personalization and Model Heterogeneity}
Heterogeneity
FedLoRA~\cite{yi2023fedlora} designs a homogeneous adapter to accommodate federated clients’ heterogeneous local model training, which can be aggregated on the fine-tuning server.
HETLORA~\cite{cho2024heterogeneous} instead uses heterogeneous LoRA designs across devices to reduce training imbalanced problems, and sparsity-weighted aggregation is developed on the server.
FlexLoRA~\cite{bai2024federated} also enables heterogeneous LoRA ranks, which are aggregated on the server using Singular Value Decomposition. FedPCL~\cite{tan2022federated} is proposed to federally fine-tune client-specific and class-relevant information with homogeneous local models as prototypes. By prototype-wise contrastive learning, the aggregated model can preserve personalized model representation while achieving similar performance with a large-scale centrally trained model.
FedBRB~\cite{xu2024fedbrb} also proposes a small-to-large FL framework, where small local models are fine-tuned as blocks of a large global model.

\subsection{Privacy and Security}

\textbf{Security and Robustness}~\cite{lu2023exploring} comprehensively discussed potential vulnerabilities and essential components in decentralized training for LLMs in terms of data and models.~\cite{zhao2023privacy} studied a comprehensive literature survey about privacy-preserving in federated training of foundation models.~\cite{wu2024vulnerabilities} investigated the vulnerability of federated foundation models under adversarial threats. In addition, a novel attack strategy is developed to target the safety issues of foundation models in the setting of FL.

\noindent\textbf{Privacy}
To enhance privacy protection and improve model performance, FedPIT~\cite{zhang2024fedpit} proposed to use LLMs' in-context learning abilities to augment local synthetic datasets.  
FFA-LoRA~\cite{sun2024improving} fixes the randomly initialized non-zero matrices and only fine-tunes the zero-initialized matrices in LoRA to solve learning inefficiency in privacy-preserved FL.

\noindent\textbf{Attacks}
Several papers also discuss the potential adversarial attack strategies and situations in federated LLMs fine-tuning.
Wu et al. ~\cite{wu2024vulnerabilities} investigated the vulnerability in federated fine-tuning under adversarial threats. 
Experiments are conducted in both image and text domains, where this work discovered the high susceptibility of federated fine-tuning to adversarial attacks.
Attackers can also recover private text in federated learning of language models~\cite{gupta2022recovering} with high fidelity.
Further experiments show that if in the fine-tuning stage, the pre-trained language model's word embedding remains unchanged, such attacks can be less effective.
Decepticon~\cite{fowl2022decepticons} also proposes a latent vector attack, which can discover private user text with malicious parameter vectors.
In heterogeneous federated learning, Li et al. ~\cite{li2023unveiling} also develop a novel backdoor attack mechanism by mimicking normal client behavior.

\noindent\textbf{Defense}
Defense for privacy-preserving and malicious attacks can be beneficial and challenging for LLMs, due to the efficiency-effectiveness balance for large-scale models.
FEDML-HE~\cite{jin2023fedml} proposes a homomorphic encryption algorithm that selectively encrypts sensitive parameters and balances security and efficiency.
The optimized federated fine-tuning method can significantly reduce the computation overhead. \cite{yu2023federated} more comprehensively studied the federated foundation model in privacy-preserving pre-training, fine-tuning, and application.
In addition, intellectual property protection is also crucial in federated LLM fine-tuning, where FedBiOT~\cite{wu2023fedbiot} creates a bi-level optimization problem to ensure the emulator, trained on a public dataset by the LLM owner supporting adaptors in fine-tuning on clients' private datasets, which works effectively regardless of distribution drift between the datasets.
\subsection{Efficiency}

\textbf{Training Efficiency}
In LLM-based federated learning, optimizing training efficiency is crucial for minimizing latency and computational overhead. 
This ensures rapid and effective fine-tuning across diverse and distributed private datasets.
Dataset Grouper~\cite{charles2024towards}, creates large-scale federated datasets to address the data efficiency problem in federated LLM fine-tuning, 
which support flexible and scalable federated learning simulations for training language models with billions of parameters.
FedYolo~\cite{zhang2023fedyolo} explores possibilities in on-device training of large language models,
emphasizing the benefits of model scale and modularity to enhance heterogeneity robustness, 
reduce communication costs, and prevent catastrophic forgetting while enabling clients to adapt to diverse tasks with a single general-purpose model.
FedTune~\cite{chen2022fedtune} benchmarks three fine-tuning methods (e.g., modifying the input, adding extra modules, and adjusting the backbone) and conducts empirical experiments to show the comprehensive comparison between federated learning and ordinary fine-tuning.

\noindent\textbf{Communication Efficiency}
Managing communication costs while ensuring data privacy is paramount in federated fine-tuning for large language models. 
Recent research has explored innovative solutions to reduce communication overhead while maintaining or enhancing the performance of federated systems.
The integration of federated learning in HRI and robots benefits from data privacy but suffers from communication inefficiency.
Khalid et al.~\cite{khalid2023cefhri} propose a communication-efficient FL framework for HRI (CEFHRI), leveraging pre-trained models and a trainable spatiotemporal adapter for video understanding tasks.
FedKSeed~\cite{qin2023federated}, explores federated full-parameter fine-tuning, which can be communication inefficient in LLMs, which utilizes zeroth-order optimization with finite random seeds, significantly reducing transmission requirements to a few thousand bytes. 
FedRDMA~\cite{zhang2024fedrdma}, a cross-silo FL system, integrates RDMA into the FL communication protocol to enhance efficiency by dividing the updated model into chunks and optimizing RDMA-based communication.

\noindent\textbf{Parameter Efficiency}
Several existing papers have explored parameter-efficient learning methods for fine-tuning large language models (LLMs) in federated learning (FL)~\cite{wu2024fedbiot}. These methods address the communication, computation, and storage constraints of edge devices, making federated training more efficient.~\cite{sun2022exploring} systematically investigates ways to reduce communication in federated fine-tuning, while~\cite{zhang2023fedpetuning} study four PETuning methods and propose a benchmark to defend against privacy attacks.~\cite{wang2024personalized} introduce a gradient-free prompt tuning method for better performance in few-shot settings. In heterogeneous FL environments, several LoRA-based methods have been developed: SLoRA~\cite{babakniya2023slora} enables fine-tuning with sparse parameter masks and SVD-decomposed LoRA blocks, FFA-LoRA~\cite{sun2023improving} mitigates LoRA instability due to data heterogeneity, and FlexLoRA~\cite{bai2024federated} adjusts LoRA ranks dynamically. LP-FL~\cite{jiang2023low} leverages few-shot prompt learning and soft labeling for efficient communication.

\subsection{Frameworks}
\textbf{Cross-Silo}
Federated learning (FL) in cross-silo settings presents unique challenges, particularly in communication efficiency and collaborative learning between different model sizes. 
FedRDMA~\cite{zhang2024fedrdma} is a communication-efficient FL system integrating Remote Direct Memory Access (RDMA) to alleviate the communication overheads exacerbated by LLMs. 
FedRDMA significantly improves communication efficiency over traditional TCP/IP-based systems by segmenting the updated model and employing optimization techniques. 
CROSSLM~\cite{deng2023mutual} uses smaller language models (SLMs) trained on private task-specific data to alleviate privacy concerns, which can further augment the learning of LLMs. 
This mutual enhancement approach allows both SLMs and LLMs to improve their performance while preserving LLMs' generalization capabilities.

\noindent\textbf{Cross-Device}
Different from cross-silo settings, normally, cross-device federated learning involves a large number of small-scale end devices, where fine-tuning LLMs can be challenging. 
A systematic study~\cite{wang2023can} has used large-scale public data and LLMs to explore distillation methods in training large-scale parallel devices to improve privacy-utility tradeoff.
To address the challenge of resource constraints on mobile devices, FwdLLM~\cite{xu2023federated} proposes a backpropagation-free training protocol that enhances memory and time efficiency in federated learning of LLMs, facilitating the deployment of billion-parameter LLMs on commercial mobile devices.
Woisetschläger et al.~\cite{woisetschlager2023federated} focus on energy efficiency in federated learning of LLMs on edge devices. 
The FLAN-T5 foundation models are evaluated end-to-end for fine-tuning with a real-time metric of energy efficiency to assess the computational efficiency.

\noindent\textbf{Decentralized Training}
The extensive computational resources required for training and deploying LLMs have become a significant challenge. 
Recent research has explored decentralized FL as a solution, leveraging underutilized consumer-level GPUs and private data to make LLM training more accessible and privacy-preserving.
OpenFedLLM~\cite{ye2024openfedllm} proposes a complete pipeline for training LLMs on the underutilized distributed private data via FL. 
Specifically, to leverage consumer-level GPUs for LLM training, Tang et al.~\cite{tang2023fusionai} proposed a decentralized system that includes a dynamic broker with a backup pool to handle the variability and heterogeneity of devices. 
Intermediate representation and execution planes are abstracted to ensure compatibility across different devices and deep learning frameworks.
Yuan et al.~\cite{yuan2022decentralized} propose a scheduling algorithm that efficiently allocates computational tasks across decentralized GPUs connected by slow networks, which enables decentralized training among heterogeneous and lower-bandwidth interconnected devices.

\noindent\textbf{Blackbox and Transfer Learning}
When dealing with black-box foundation models, gradients are often unavailable, 
where zeroth-order optimization techniques are explored as a means to update models without requiring gradient access.
Fed-BBPL ~\cite{lin2023efficient} optimizes prompt generators in federated learning settings without directly accessing the foundation models' parameters for text and image classification tasks.
ZooPFL ~\cite{lu2023zoopfl} presents a more general approach, making it suitable for more complex and diverse FL tasks.
\cite{kang2023grounding} instead focus on transfer learning that enables clients to benefit from general knowledge learned in federated settings and also transfer domain-specific knowledge back to improve the federated learning.

\noindent\textbf{Instruction Tuning}
FederatedScope-LLM~\cite{kuang2023federatedscope} includes multiple efficient federated fine-tuning algorithms to reduce communication and computation costs,
where clients may or may not have access to the full model parameters.
FedIT ~\cite{zhang2023towards} designs FL approaches to incorporate diverse user instructions on local devices, which might be data sensitive.

\subsection{Evaluation}

\textbf{Datasets}
FederatedScope-LLM~\cite{kuang2023federatedscope} is a comprehensive fine-tuning package for diverse federated settings, including various training and evaluation datasets.
Communication and computation costs, as well as memory and multi-GPU efficiency, are considered.
OpenFedLLM~\cite{ye2024openfedllm} is another federated LLM training framework, where clients are trained on underutilized distributed private data and can collaboratively train a shared
model without data transmission.

\noindent\textbf{Benchmarks}
FedLLM-Bench \cite{ye2024fedllm} focuses on providing user-annotated datasets in several settings, including multilingual collaborative learning, multi-user preferences, and conversations from different chatbots. 
Fedmlsecurity~\cite{han2023fedmlsecurity}, on the other hand, focuses more on the attacks and defense scenarios in federated learning. 
This benchmark provides various backbone model implementations and different attacks and defense configurations.
Profit~\cite{collins2023profit} benchmarks the trade-off between personalization and robustness in federated learning. 
The benchmarking results show robust performance when using a small learning rate with many local epochs for personalization, especially with an adaptive client-side optimizer.

\noindent\textbf{Convergence Analysis}
FedPEAT~\cite{chua2023fedpeat} proposes a server-device collaborative learning framework, where parameter-efficient learning methods and an emulator are incorporated. 
Convergence analysis in FedPEAT is comprehensive to demonstrate that it's 4.6 times faster than the conventional Fed-FT and uses 4.9 times less communications compared with it.
FedMeZO~\cite{ling2024convergence} integrates a zeroth-order optimization method in federated LLM learning, which converges faster than conventional methods and also significantly reduces GPU memory usage.

\section{Prompt Learning}
Prompt-based learning involves creating a prompting function to achieve optimal performance on a task \cite{liu23survey}. This requires either a human or an algorithm to choose the best template for each task. In FL, large LLMs incur high communication costs and need significant local data for training. Prompt learning, which fine-tunes soft prompts without altering LLMs, emerges as a promising method to reduce communication costs, as shown in Table~\ref{tab:topic_prompt_learning}.

\begin{table*}[ht]
\caption{Overview of Topics in Prompt Learning.}
\vspace{-1em}
\centering
{
\begin{tabular}{|c|c|}
\hline

\textbf{Topics}                  & \textbf{Approaches}                                                                                                                                                                                                                                                                                                                                                       \\ \hline
Prompt Generation       & FedTPG~\cite{qiu2023text}, TPFL~\cite{zhao2023inclusive}                                                                                                                                                                                                                                                                   \\ \hline
Few-Shot Scenario       & FedFSL~\cite{cai2023federated}                                                                                                                                                                                                                                                                                                                   \\ \hline
Chain of Thoughts       & Fed-SP-SC~\cite{liu2023federated}, FedLogic~\cite{xing2023fedlogic}                                                                                                                                                                                                            \\ \hline
Personalization         & pFL~\cite{guo2023pfedprompt}, FedLogic~\cite{yang2023efficient}, Fed-DPT ~\cite{wei2023dual}                                                                                                                                                                                                   \\ \hline
Multi-Domain            & FedAPT~\cite{su2022federated}, Profit~\cite{collins2023profit}                                                                                                                                                                                                                                                                   \\ \hline
Parameter Efficient     & FedPepTAO~\cite{che2023federated}, FedLoRA~\cite{wu2023fedlora}                                                                                                                                                                                                                                                      \\ \hline
Communication Efficient & FedPrompt~\cite{zhao2023fedprompt}                                                                                                                                                                                                                                                                                                         \\ \hline
Blackbox                & FedBPT~\cite{sun2023fedbpt}                                                                                                                                                    \\ \hline
Retrieval Augmented Generation                & FeB4RAG~\cite{wang2024feb4rag}                                                                                                                                                                                         \\ \hline
Applications            & \begin{tabular}[c]{@{}c@{}}Multilingual~\cite{zhao2023breaking}, Recommender Systems~\cite{guo2024prompt,zhao2024llm}, \\ Medical VQA~\cite{zhu2024prompt}, Weather Forecasting~\cite{chen2023prompt},\\ Virtual Reality~\cite{zhou2024federated}\end{tabular} \\ \hline
\end{tabular}
}
\label{tab:topic_prompt_learning}
\end{table*}



\subsection{Prompt Generation}
PromptFL~\cite{guo2023promptfl} emphasizes training prompts instead of a shared model, using foundation models like CLIP for efficient global aggregation and local training, even with limited data. This boosts privacy, efficiency, and performance in FL environments. ~\cite{qiu2023text} introduces a context-aware prompt generation network based on task-related textual inputs, improving model adaptability to diverse classes across remote clients. ~\cite{zhao2023inclusive} presents Twin Prompt Federated Learning (TPFL), which integrates visual and textual modalities to overcome the limitations of single-modality prompt tuning in federated learning, improving data representation across nodes.


\subsection{Few-shot Scenario}


LLMs are fine-tuned within a federated, privacy-preserving framework for various language tasks, relying on hundreds of thousands of labeled samples from mobile clients. However, mobile users often resist or lack expertise in data labeling, leading to a \textbf{few-shot scenario}.~\cite{cai2023federated} introduce FeS, the first framework for Federated Few-Shot Learning in mobile NLP. FeS incorporates curriculum pacing with pseudo labels, representational diversity for sample selection, and co-planning of model layer depth and capacity. This reduces training time, energy, and network traffic, enabling efficient fine-tuning on resource-limited devices.


\subsection{Chain of Thoughts}
Recent research shows that Chain of Thought (CoT), a series of reasoning steps, greatly improves LLMs’ complex reasoning abilities~\cite{jason2022cot}. Fed-SP-SC~\cite{liu2023federated} improve LLM accuracy by using federated prompting with CoT reasoning, which employs a majority voting system on synonymous questions to boost zero-shot reasoning. FedLogic \cite{xing2023fedlogic} enhances LLM interpretability and effectiveness through optimized CoT prompt selection across domains, using variational expectation maximization (V-EM) for cross-domain personalization. These advancements improve LLM adaptability and reasoning capabilities.

 
\subsection{Personalization}
Inter-client uncertainty refers to differences between users’ local models within the same time frame. Higher uncertainty signals more diverse local data distributions among devices, emphasizing the need for personalization.~\cite{guo2023pfedprompt} introduce a prompt-based personalized FL (pFL) method for medical VQA, addressing data heterogeneity and privacy by treating distinct organ datasets as separate clients. This method uses pFL to develop unique transformer-based VQA models per client while reducing the communication burden via compact prompt parameters. A reliability parameter prevents performance loss and irrelevant contributions.~\cite{yang2023efficient} propose FedLogic, integrating FL with prompt selection optimization for LLMs to enhance chain-of-thought (CoT) reasoning across domains while maintaining privacy. FedLogic balances generality and personalization in CoT prompts by exchanging knowledge between models, fine-tuning rule learning, and optimizing performance through a bi-level program with KL-divergence constraints. \cite{wei2023dual} propose Fed-DPT, which leverages pre-trained visual language models and dual prompt tuning to improve federated learning in scenarios with domain-diverse client data.

\subsection{Multi-domain}
Su et al.~\cite{su2022federated} introduce FedAPT, which integrates FL with adaptive prompt tuning to tackle multi-domain collaborative learning. FedAPT personalizes prompts for each client’s domain while enabling collaboration across clients without sharing raw data. Experiments show it outperforms existing methods in handling domain shifts.~\cite{collins2023profit} address the gap by evaluating FedAvg and FedSGD with personalization in the context of prompt tuning for LLMs. Their benchmark highlights the robustness of federated-trained prompts with small learning rates and many local epochs, showing that regularization and prompt interpolation improve the personalization-robustness trade-off in limited computation settings.

\subsection{Parameter Efficient}
The training process of LLMs generally incurs the update of significant parameters, which limits the applicability of FL techniques to tackle the LLMs in real scenarios.
Prompt tuning can significantly reduce the number of parameters to update, but it either incurs performance degradation or low training efficiency.

\noindent\textbf{Prompt Tuning}~\cite{che2023federated} puts forth a novel Parameter-efficient prompt Tuning approach, FedPepTAO, encompassing Adaptive Optimization to facilitate efficient, effective FL of LLMs. Firstly, a unique and efficient partial prompt tuning method is proposed to simultaneously optimize both performance and efficiency. Secondly, a novel adaptive optimization technique is introduced to combat client drift issues across both the device and server sides and thus enhance overall performance.

\noindent\textbf{LowRank Adaptation}
Inspired by recent advancements in fine-tuning LLMs, particularly the LowRank Adaptation (LoRA) technique, which preserves general linguistic knowledge in a pre-trained full-rank model while retaining domain-specific knowledge in a low-rank matrix,~\cite{wu2023fedlora} propose FedLoRA for PFL. FedLoRA aims to maintain shared general knowledge in a common full-rank matrix and capture client-specific knowledge in a personalized low-rank matrix. However, unlike LoRA, FedLoRA's full-rank matrix requires training from scratch, which is affected by data heterogeneity. To address this, they propose a new training strategy to mitigate these effects.

\subsection{Communication Efficient}

FedPrompt~\cite{zhao2023fedprompt} combines FL with prompt tuning to tackle challenges in decentralized data processing, prioritizing communication efficiency and privacy. Their method uses a split aggregation framework, significantly reducing communication costs by transmitting only 0.01\% of a pre-trained language model's parameters, with minimal accuracy loss on both IID and Non-IID data. FedPrompt enhances FL efficiency while maintaining strong privacy protections, making large language models more practical for decentralized applications.



\subsection{Blackbox}

Applying FL to black-box LLMs faces significant challenges, including restricted access to model parameters, high computational demands, and substantial communication overheads. To address these issues, Sun et al. \cite{sun2023fedbpt} introduce Federated Black-box Prompt Tuning (FedBPT), an innovative framework that overcomes these obstacles. FedBPT's key advantage lies in its ability to operate without client access to model parameters. The framework focuses on training optimal prompts and employs gradient-free optimization methods, effectively reducing the number of exchanged variables. This approach significantly enhances communication efficiency and minimizes both computational and storage costs. 

\subsection{Applications}
Researchers have explored diverse applications of FL in LLM-based systems across domains such as multilingual processing, recommender systems, medicine, weather forecasting, and virtual reality. These studies illustrate how FL can be integrated with techniques including prompt tuning, personalization, and domain-specific adaptations to improve model performance while preserving data privacy and addressing challenges inherent to distributed environments.

\noindent\textbf{Multilingual}: Zhao et al. \cite{zhao2023breaking} propose Multilingual Federated Prompt Tuning to enable parameter-efficient adaptation of multilingual LLMs without sharing raw data. Local prompt encoders capture region-specific linguistic characteristics, and aggregated updates produce a global model that better generalizes to low-resource languages while respecting data restrictions.

\noindent\textbf{Recommender Systems}: Guo et al. \cite{guo2024prompt} introduce the Privacy-Preserving Federated Content Representation (PFCR) framework, which combines FL with secure gradient encryption and prompt-based content representation. By leveraging textual item descriptions and shared feature spaces, PFCR enhances cross-domain recommendation while mitigating privacy risks.

\noindent\textbf{Medical VQA}: Zhu et al. \cite{zhu2024prompt} present a personalized prompt-based FL approach for medical VQA, treating datasets from different organs as separate clients. Lightweight prompts enable efficient knowledge sharing, while a reliability weighting mechanism reduces negative influence from low-quality clients.

\noindent\textbf{Weather Forecasting}: Chen et al. \cite{chen2023prompt} develop a federated spatio-temporal foundation model for cross-regional weather forecasting. Prompt learning improves adaptation to heterogeneous and low-resource sensors, enabling collaborative forecasting without compromising sensitive meteorological data.

\noindent\textbf{Virtual Reality}: FedPromptDT~\cite{zhou2024federated} integrates FL and prompt-based sequence modeling for personalized resource allocation in mobile edge VR services. Prompt design allows rapid adaptation to varying environments and user preferences without retraining, improving efficiency and user experience.




\section{Emerging Directions} 

While significant strides have been made in the integration of federated learning with LLMs as discussed above, several promising avenues remain underexplored. 
In this section, we discuss potential directions for future work shown in Table~\ref{tab:topic_potential_directions}.

\begin{table*}[ht]
\caption{Overview of Topics in Potential Directions}
\vspace{-1em}
\centering
{
\begin{tabular}{|c|c|}
\hline
\textbf{Area}                           & \textbf{Topic}                       \\ \hline
\multirow{2}{*}{Real World Deployment}  & Personalized FL on Confidential Data \\ \cline{2-2} 
                                        & Collaborative FL                     \\ \hline
Multi-Modality Model                    & Modality Co-optimization             \\ \hline
\multirow{2}{*}{Federated Pre-Training} & Efficient Data Exchange              \\ \cline{2-2} 
                                        & Model Architecture Design            \\ \hline
\multirow{2}{*}{Federated AI Agents}                     & Federated Inference on Local Data        \\ \cline{2-2} 
                                        & Federated Agents across Clients
                                        \\ \hline
\multirow{3}{*}{LLMs for FL}            & Synthetic FL Data Generation         \\ \cline{2-2} 
                                        & Capacity-Augmented FL                \\ \cline{2-2} 
                                        & Responsible and Ethical LLM4FL       \\ \hline
\end{tabular}
}
\label{tab:topic_potential_directions}
\end{table*}


\subsection{Real World Deployment} 
One of the critical challenges in deploying federated LLMs to ensuring that these models can operate effectively in real-world scenarios, where the data is often heterogeneous and the computing resources are often constrained. 
Recent works have introduced techniques like LoRA within personalized federated learning to address these challenges by maintaining shared general knowledge while capturing client-specific adaptations, mitigating the impact of data heterogeneity~\cite{wu2023fedlora}.

However, despite these advances, optimizing federated learning frameworks for deployment in more diverse real-world environments remains a challenge. 
For example, developing personalized AI agents that can fine-tune on confidential data while respecting user privacy will require innovations in privacy-preserving techniques to enable collaborative learning with large-scale models.
Additionally, while efficient model adaptation will be critical in enhancing the performance and scalability of LLMs in decentralized settings \cite{ye2024openfedllm}, current approaches often make trade-offs between model accuracy and resource efficiency.
Furthermore, robust deployment strategies that can handle the variability in real-world data and infrastructure, particularly in resource-constrained environments, may also benefit from techniques such as swarm intelligence \cite{qu2024federated} that distribute computational tasks across multiple agents.


\subsection{Multimodality Models} 

Integrating multimodal data into federated LLMs presents opportunity for enhancing the versatility and applicability of these models. 
Methods such as FedClip~\cite{lu2023fedclip} have shown potential in combining visual and textual data to improve performance across various tasks.
However, how to ensuring seamless integration of different data modalities while maintaining data privacy and model efficiency poses new challenges.
Recent studies, such as \cite{xu2024survey} have noted the complexity of aligning diverse data modalities in decentralized settings, which often results in sub-optimal performance and increased computational costs.
Future research may explore the co-optimization of multiple models handling different modalities, focusing on reducing these inefficiencies and developing robust frameworks for comprehensive and context-aware AI systems in federated environments.

\subsection{Federated Pre-Training} 

Given the high computational costs associated with pre-training LLMs, there is a growing need for strategies that reduce these costs without sacrificing model performance.
Federated pre-training \cite{tian2022fedbert,wen2024pre} offers a potential solution by enabling models to be adapted to specific domains without relying on extensive centralized datasets.
Among the challenges in federated pre-training, the efficiency of data exchange protocols remains a crucial factor.
Recent strategies like fully-sharded data parallelism \cite{sani2024future}, which aims to minimize memory consumption and optimize communication between GPUs during training, have made federated pre-training more viable even with limited hardware resources.
Future research may further explore these techniques to strike a better balance between computational costs and the benefits of federated pre-training, possibly by optimizing data exchange protocols or developing more efficient model architectures.


\subsection{Federated AI Agents}
\textbf{Federated Inference on Local Data} One promising direction is the development of techniques for federated inference, where models trained in a federated manner are used to make predictions directly on local data. 
This approach would help mitigate privacy concerns while allowing for real-time, on-device inference. 
However, optimizing the inference process to minimize latency and computational overhead remains challenging.
Prior works such as \cite{liu2022federated} adapt pre-trained models for federated contexts during inference using parameter-efficient learning to minimize the computational load.
\cite{ma2022federated} has also explored aligning local representations with global models to enhance the accuracy and responsiveness in federated inference.
While these works focus on smaller models, future work may develop new federated inference techniques for LLMs, addressing the trade-offs between model complexity, real-time performance, and resource efficiency, which are significant for large-scale models.

\noindent\textbf{Federated Agents across Clients} Beyond on-device inference, an emerging direction is to treat each client as an \emph{agent}, an LLM-driven entity that observes local environments, calls tools, and participates in federated optimization. FedLLM systems evolve into distributed populations of heterogeneous agents with local capabilities (e.g., private RAG or workflow execution) that align toward shared global objectives under privacy and resource constraints~\cite{jing2025federated,krouka2021communication}.

Recent work explicitly explores \emph{LLM agents under FL}. FICAL~\cite{wu2024federated} enables communication-efficient coordination by exchanging compact knowledge compendiums instead of parameters, while also teaching agents tool-use via RAG modules. FedAgent~\cite{sun2025fedagent} demonstrates that decentralized RL training can optimize agent policies across clients without sharing trajectories, achieving near-centralized performance while preserving data locality.

FMARL research provides key algorithmic foundations, such as federated multi-agent deep RL~\cite{jing2025federated} and heterogeneity-aware approaches like FedMRL~\cite{sahoo2024fedmrl}, which support client-specific policies and robust aggregation across non-IID environments. These ideas transfer naturally to FedLLM agents that must collaborate across organizations with divergent goals.

At the systems level, frameworks such as Academy~\cite{pauloski2025empowering} deploy stateful LLM-based agents across federated research infrastructure, enabling distributed execution while preserving institutional autonomy. In federated RAG settings, each client acts as a local retrieval agent that queries its private corpus and contributes only minimal retrieval signals (top-k results or relevance scores) to the global generation process, avoiding raw data exchange~\cite{wang2024feb4rag}.

Open challenges remain in coordination and credit assignment, communication efficiency, personalization with safety, and standardized evaluation. Addressing these issues will be crucial for transforming FedLLM from distributed model training into large-scale, privacy-preserving multi-agent systems.

\subsection{LLMs for Federated Learning}

\paragraph{Synthetic FL Data Generation} As we discussed above in Section~\ref{sec:intro}, concerns about data privacy and data scarcity can severely limit the effectiveness of models in FL scenarios ~\cite{FedAffect,gao2022survey,chen2022practical}, especially when sensitive or limited data are involved, such as in the healthcare or financial sectors \cite{liu2023efficient,zhang2022dense}. Large Language Models propose a promising solution by generating synthetic data, which can significantly enrich the original training sets \cite{wang2022self,li2023synthetic,wei2023simple}. This synthetic data, being reflective of diverse real-world scenarios, helps alleviate the issue of data scarcity and introduces additional diversity, thus enhancing the model’s generalization capabilities and reducing the likelihood of overfitting \cite{goetz2020federated,dankar2022using,wu2024prompt}.

\paragraph{Capacity-Augmented FL} Furthermore, LLMs offer a solid foundation for improving FL performance \cite{yu2023federated,zhang2024mllm} through advanced capabilities such as knowledge distillation \cite{li2019fedmd,yang2023knowledge} and prompt engineering \cite{zhao2023fedprompt}. These models provide a starting point for local fine-tuning on clients’ data, which enables faster convergence and potentially higher model accuracy without the extensive need for ground-up model training \cite{nguyen2022begin,chen2022importance}. In addition, by leveraging the rich data understanding inherent in LLMs, suboptimal performance due to limited or imbalanced data can be substantially mitigated, thus improving efficiency and effectiveness across various FL deployments \cite{zhang2024mllm}.

\paragraph{Responsible and Ethical LLM4FL} However, the integration of LLMs into FL also brings forth complex challenges concerning law compliance and responsible technology usage \cite{wachter2024large,darji2024challenges,zhong2020does,wang2023where}. The generation of synthetic data by LLMs, while beneficial, must be carefully managed to avoid potential privacy infringements and intellectual property violations. Ensuring that synthetic data does not too closely resemble the original training sets is crucial to mitigate the risks of revealing sensitive information. These legal and ethical considerations necessitate a framework that not only advances the technological capabilities of FL but also aligns with stringent regulatory standards to ensure the responsible deployment of these powerful models in sensitive domains.

\newpage
\bibliographystyle{plainnat}
\bibliography{ref}

\end{document}